\documentclass[11pt,a4paper]{article}
\usepackage{times,latexsym}
\usepackage{url}
\usepackage[T1]{fontenc}
\usepackage{graphicx}
\usepackage{booktabs}
\usepackage{adjustbox}
\usepackage{url}  
\usepackage[hidelinks]{hyperref} 
\usepackage[acceptedWithA]{tacl2021v1}
\usepackage{xspace,mfirstuc,tabulary}

\newif\iftaclinstructions
\taclinstructionsfalse 
\iftaclinstructions
\renewcommand{\confidential}{}
\renewcommand{\anonsubtext}{(No author info supplied here, for consistency with
TACL-submission anonymization requirements)}
\newcommand{\instr}
\fi

\iftaclpubformat 

\else

\fi

\title{Llama-GENBA-10B: A Trilingual Large Language Model for German, English and Bavarian}

\begingroup
\author{
  Michael Hoffmann\textsuperscript{*} \\
  \small Leibniz Supercomputing Centre (LRZ) \\
  \small Garching, Germany \\
  \small \nolinkurl{Michael.Hoffmann@lrz.de}
  \And
  Jophin John\textsuperscript{*} \\
  \small Leibniz Supercomputing Centre (LRZ) \\
  \small Garching, Germany \\
  \small \nolinkurl{Jophin.John@lrz.de}
  \AND
  Stefan Schweter \\
  \small Independent Researcher \\
  \small Holzkirchen, Germany \\
  \small \nolinkurl{Stefan@Schweter.bayern}
  \And
  Gokul Ramakrishnan \\
  \small Cerebras Systems \\
  \small Sunnyvale, USA \\
  \small \nolinkurl{gokul.ramakrishnan@}\\[-0.2ex]\small \nolinkurl{cerebras.net}
  \And
  Hoi-Fong Mak \\
  \small Leibniz Supercomputing Centre (LRZ) \\
  \small Garching, Germany \\
  \small \nolinkurl{Hoi-fong.Mak@lrz.de}
  \AND
  Alice Zhang \\
  \small Cerebras Systems \\
  \small Sunnyvale, USA \\
  \small \nolinkurl{Alice.Zhang@cerebras.net}
  \And
  Dmitry Gaynullin \\
  \small Cerebras Systems \\
  \small Sunnyvale, USA \\
  \small \nolinkurl{dmitry.gaynullin@}\\[-0.2ex]\small
  \nolinkurl{cerebras.net}
  \And
  Nicolay J. Hammer \\
  \small Leibniz Supercomputing Centre (LRZ) \\
  \small Garching, Germany \\
  \small \nolinkurl{Nicolay.Hammer@lrz.de}
}

\endgroup

\date{}

\begin{document}
\maketitle
\begingroup
  \renewcommand{\thefootnote}{\fnsymbol{footnote}}
  \footnotetext[1]{Equal contribution.}
\endgroup

\begin{abstract}
We present Llama-GENBA-10B, a trilingual foundation model addressing English-centric bias in large language models. Built on Llama 3.1-8B and scaled to 10B parameters, Llama-GENBA-10B is continuously pretrained on 164B tokens (82B English, 82B German, and 80M Bavarian), balancing resources while preventing English dominance. Targeted at the German NLP community, the model also promotes Bavarian as a low-resource language. Development tackled four challenges: (1) curating a multilingual corpus despite Bavarian scarcity, (2) creating a unified tokenizer for English, German, and Bavarian, (3) optimizing architecture and language-ratio hyperparameters for cross-lingual transfer, and (4) establishing the first standardized trilingual evaluation suite by translating German benchmarks into Bavarian. Evaluations show that Llama-GENBA-10B achieves strong cross-lingual performance, with the fine-tuned variant surpassing Apertus-8B-2509 and gemma-2-9b in Bavarian and establishing itself as the best model in its class for this language, while also outperforming EuroLLM in English and matching its results in German. Training on the Cerebras CS-2 demonstrated efficient large-scale multilingual pretraining with documented energy use, offering a blueprint for inclusive foundation models that integrate low-resource languages.

\end{abstract}


\section{Introduction}
Large Language Models (LLMs) have evolved from English-dominant architectures to embrace multilingual capabilities, spanning proprietary systems like GPT-4 \cite{achiam2023gpt}, Gemini 2 \cite{team2023gemini}, and Qwen-3 \cite{yang2025qwen3}, alongside open-source alternatives such as Llama 2 \cite{touvron2023llama} and Llama 3 \cite{dubey2024llama}. Recent advances have expanded language coverage to over 100 languages through models like xLlama-100 and xBLOOM-100 \cite{lai2024llms}, yet substantial challenges remain in achieving balanced representation across diverse linguistic communities.

Despite these developments, English continues to dominate pretraining datasets and evaluation frameworks. Leading open-source models like Llama-3 \cite{dubey2024llama}, while supporting multilinguality across dozens of languages, remain predominantly trained on English-centric corpora, treating multilingual performance as secondary rather than foundational. Although truly multilingual architectures trained on more balanced data—such as Bloom \cite{workshop2022bloom} and PolyLM \cite{wei2023polylm}—are emerging, they remain the exception. Consequently, performance disparities persist, particularly affecting underrepresented languages \cite{wang2020negative} and highlighting the critical need for models that prioritize multilingual parity from inception.

This study introduces GENBA-10B, a trilingual model for German, English, and Bavarian, with less than half of the training data in English. The model is intended to support the German and Bavarian NLP community, and enable applications ranging from tourism to media accessibility while contributing to linguistic preservation\footnote{Initial distribution is restricted to approved academic and non-profit organizations under a non-commercial license; we are working to resolve the legal aspects for broader release.}.

Built upon Llama3.1-8B and expanded to 10B parameters through block expansion techniques \cite{gosal2024bilingual,wu2024llama}, GENBA-10B was trained on 164B tokens comprising 82B English, 82B German, and 80M Bavarian tokens. Development addressed four critical challenges: curating training data despite the scarcity of Bavarian resources; developing a custom trilingual tokenizer; selecting optimal architecture and language distribution ratios; and establishing appropriate evaluation benchmarks, including Bavarian-specific assessment tools.

We evaluate GENBA-10B with a trilingual benchmark suite translating German tasks (HellaSwag, ARC, TruthfulQA, MMLU) into Bavarian, enabling direct cross-lingual comparison across Bavarian, German, and English. In its base form, GENBA-10B ranks second among European models, following Apertus-8B. It performs particularly well in English and Bavarian, and achieves comparable results to EuroLLM in German. After trilingual fine-tuning, the model attains state-of-the-art performance in Bavarian, even surpassing Apertus-8B and Gemma-2-9B-it, while also outperforming EuroLLM in English and matching its results in German. Overall, the fine-tuned variant establishes GENBA-10B as the strongest sub-10B parameter model for Bavarian language tasks.

Training on a single  Cerebras CS-2 AI Accelerator \cite{cerebrasc2system} shows that large-scale multilingual pretraining can be conducted efficiently by small research teams\footnote{GENBA-10B was developed by a team of five core contributors.}. We tracked energy consumption and training time to evaluate hardware efficiency and resource use, providing practical insights into the computational costs of foundation model development and guidance for resource-constrained groups.

This study contributes to multilingual Large Language Model development in three ways:

First, we present GENBA-10B, a trilingual model designed to mitigate data scarcity, tokenizer limitations, cross-lingual transfer challenges, and English dominance by training on a corpus with equal-scale English and German components (82B tokens each) and a complementary Bavarian component (80M tokens).

Second, we establish the first trilingual benchmark suite and show that GENBA-10B achieves strong cross-lingual performance, with the fine-tuned variant setting a new state of the art for Bavarian among sub-10B models.   

Third, we demonstrate efficient large-scale pretraining on a single Cerebras CS-2 system while recording energy use, providing practical guidance for resource-constrained teams.

The remainder of this study is organized as follows: Section \ref{RelatedWork} reviews related work on multilingual and dialect-specific LLMs; Section \ref{Approach} outlines our methodology, including data collection, tokenizer design, architecture, and training infrastructure; Section \ref{Experiments} reports experiments on language ratios, architecture scaling, hyperparameters, fine-tuning, and energy use; Section \ref{Results}  presents evaluation results for both base and fine-tuned models together with pretraining energy measurements; and Section \ref{Conclusion} concludes with future directions.

\section{Related Work}
\label{RelatedWork}
This section reviews prior research on large language models (LLMs) in two areas: multilingual LLM development as well as German-and dialect‐tailored approaches.

\subsection{Multilingual LLMs}
Transformer-based large language models (LLMs) have recently moved from English-only training to genuinely multilingual settings. Prominent examples include Llama-3.1 \cite{dubey2024llama}, Mistral 7B \cite{jiang2023mistral}, and Qwen3-8B \cite{yang2025qwen3}, which can process non-English input but rarely disclose the distribution of languages in their pretraining corpora. Although efforts such as Aya \cite{ustun-etal-2024-aya} and Llama-3.1 \cite{dubey2024llama} explicitly aim to broaden linguistic coverage, their training data remain dominated by English, constraining performance in underrepresented languages \cite{choudhury2025llama}. Moreover, these models are subject to the well-documented “curse of multilinguality” \cite{pfeiffer2022lifting}.

To mitigate these imbalances, several equity‐focused models have emerged. EuroLLM-9B \cite{martins2025eurollm} was trained from scratch on over 4 trillion tokens across three phases, covering all 24 official EU languages plus 11 additional ones. Despite this broad coverage, English still dominates, comprising 50 per cent of phase 1 and 32.5 per cent of phases 2 and 3, while no single non-English language exceeds 6 per cent of any phase.  Nonetheless, EuroLLM-9B achieves competitive results on multilingual benchmarks. Similarly, Teuken-7B \cite{ali2024teuken} trains from scratch on a corpus that is roughly 60 per cent non-English and employs a custom multilingual tokenizer to support all 24 EU languages, demonstrating competitive performance on European variants of ARC, HellaSwag, MMLU, and TruthfulQA. Finally, the Swiss AI initiative recently released Apertus \footnote{\href{https://github.com/swiss-ai/apertus-tech-report/blob/main/Apertus_Tech_Report.pdf}{https://github.com/swiss-ai/apertus-tech-report}}, open multilingual models trained from scratch on 15 trillion tokens spanning 1,000+ languages, including previously underrepresented Swiss German and Romansh. Despite only 40 per cent of the training data being non-English, the models remain competitive.  

Full multilingual pretraining yields strong results but at the cost of immense compute and data requirements. An alternative is \emph{continual pretraining} of established models on targeted corpora. Building on the block‐expansion method of \citet{wu2024llama} and \citet{gosal2024bilingual}, one can insert identity‐initialized Transformer blocks into a frozen base model and train only these new blocks on specialized data. In the Llama Pro variant, derived from Llama 2-7B with eight extra blocks (totaling 8.3 billion parameters), this approach used 80 billion tokens of code and mathematics to improve domain‐specific capabilities without catastrophic forgetting \cite{de2021continual}.

We extend block-expansion–based continual pretraining to the multilingual setting with our GENBA-10B model. Starting from a Llama-3.1-8B checkpoint, we expand the architecture to 10B parameters and continue training on 164B tokens of English, German, and Bavarian text. Unlike prior work focusing on code or mathematics, we target language-specific adaptation. GENBA-10B attains competitive results on multilingual benchmarks, illustrating that block expansion offers a resource-efficient approach to trilingual adaptation and a step toward more equitable multilingual LLMs.

\subsection{German and Dialect-Specific LLMs}
Research on German-centric language models has progressed from early encoder-only systems like GBERT and GELECTRA, trained on German text and achieving SoTA on classification and NER tasks \cite{chan2020german}, to more advanced models. In late 2023, LeoLM emerged as a Llama-2-based German foundation model pretrained on a high-quality German corpus, with versions at 7B and 13B parameters\footnote{\href{https://laion.ai/blog/leo-lm}{https://laion.ai/blog/leo-lm}}. Separately, the LLäMmlein series (120M and 1B decoder models), trained fully from scratch with transparency, matched or surpassed comparable models on benchmarks like SuperGLEBer 
\cite{pfister2024ll}. More recently, BübleLM-2B (2B parameters), adapted from Gemma-2-2B with a German-specific tokenizer, delivered substantial performance gains in tasks like commonsense reasoning and knowledge-based QA, outperforming both Gemma-2-2B and LLäMmlein-1B\footnote{\href{https://huggingface.co/flair/bueble-lm-2b}{https://huggingface.co/flair/bueble-lm-2b}}.

Modeling regional dialects like Bavarian introduces notable challenges: dialectal corpora are scarce, and models frequently default to standardized forms rather than retaining dialect nuances, sometimes resulting in ‘translationese’ \cite{riley2019translationese} outputs. For instance, recent datasets like MaiBaam \cite{blaschke2024maibaam} and Betthupferl \cite{blaschke2025multi} underscore both the scarcity of resources and the tendency of models to normalize dialectal inputs. In parallel, work on Galician by \citet{gamallo2024galician} demonstrates that a 1.3B parameter GPT-style model trained on a 2.1B word Galician corpus achieved promising improvements in fluency and adequacy.

Building on these insights, we introduce Llama-GENBA-10B, available in both base and instruction-tuned variants. The model is a 10B-parameter architecture initialized from the publicly available Llama-3.1-8B checkpoint\footnote{\href{https://huggingface.co/meta-llama/Llama-3.1-8B}{https://huggingface.co/meta-llama/Llama-3.1-8B}}. We perform continual pretraining on a 164B-token multilingual corpus comprising English, German, and Bavarian. Our methodological contribution lies in the timing and manner of dialectal data integration: Bavarian material, upsampled by a factor, is introduced only after 90 per cent of training has elapsed. This strategy provides a practical way to add dialectal resources to large-scale pretraining while supporting trilingual development.

\section{The Approach}
\label{Approach}
This section describes the methodology used to develop the trilingual GENBA-10B model. The process comprised four stages: (i) compiling a multilingual corpus, (ii) designing a tokenizer tailored to German–English–Bavarian, (iii) conducting architecture search experiments, and (iv) establishing a scalable training infrastructure. Each stage is described in the following subsections.

\subsection{Data Collection and Corpus Composition}
\textbf{Data Collection Methodology:}
The study began with assembling training data in three languages. While English and German resources were readily available, Bavarian data were scarce. Ultimately, two reliable sources were identified: the 
\textit{Boarische Wikipedia} corpus from \href{https://wortschatz.uni-leipzig.de/de/download/Bavarian} {the Wortschatz Uni Leipzig} project and four datasets from \href{https://opus.nlpl.eu/}{OPUS}.

To extract further Bavarian content systematically, the team initially developed a language identification script using META’s FastText classifier \cite{joulin2016bag}, trained on a balanced set of sentences from the Boarisch Wikipedia and standard German texts using high-performance computing infrastructure. However, this approach proved inadequate for reliably detecting Bavarian in unstructured or unlabeled corpora. The team subsequently adopted the GlotLID-m classifier \cite{kargaran2023glotlid}, which demonstrated higher accuracy and enabled the extraction of approximately 262,373 lines of Bavarian sentences from the Fine-Web dataset.

Several key methodological decisions were made to guide the collection process. To maintain a manageable scope, the team chose not to differentiate between dialectal variants within Bavarian. Although promising social media sources were identified, such as Facebook groups like \href{https://www.facebook.com/groups/328341027261813/}{\emph{Boarisch redn is 'in’}} (16,000 members, August 2024) and \href{https://www.facebook.com/groups/121572707986445/}{\emph{Niederbairisch für Anfänger und Runaways}} (504 members), these were excluded to avoid copyright concerns and expedite development. Similarly, the \href{https://www.zobodat.at/pdf/Akad-Bayern-Diverse_2_0001-1772.pdf}{\emph{Bayrisches Wörterbuch}} was omitted due to insufficient token volume.

\textbf{Corpus Composition:}
The training corpus comprises three partitions: English, German, and Bavarian.

\textit{English Corpus:}
The English partition is derived from the following sources:
\begin{itemize}
    \item Knowledge Pile \cite{fei2024query}: is a 188B large-scale dataset curated from web crawls (Common Crawl) using retrieval methods, primarily designed to enhance language models with high-quality knowledge-intensive text spanning diverse topics such as mathematics, biology, physics, and humanities,
    \item Cosmopedia \cite{benallal2024cosmopedia}: A dataset of synthetic textbooks, blogposts, stories, posts, and WikiHow articles generated by Mixtral-8x7B-Instruct-v0.1.
    \item \href{https://huggingface.co/datasets/Locutusque/UltraTextbooks-2.0}{UltraTextbooks-2.0}: A collection of high-quality synthetic and human-written textbooks spanning various subjects and programming languages.

    \item Proof-pile 2 \cite{azerbayev2023llemma}: A 55 billion token dataset of mathematical and scientific documents.

\end{itemize}
From these four datasets, we constructed an English partition of 82.85B tokens.

\textit{German Corpus:}
In recent years, several large-scale German corpora have emerged, including 
\href{https://german-nlp-group.github.io/projects/gc4-corpus.html}{G4Corpus}, 
\href{https://oscar-project.github.io/documentation/versions/oscar-2301/}{OSCAR}, 
or \href{https://huggingface.co/datasets/occiglot/occiglot-fineweb-v1.0}{Occiglot} 
as well as smaller collections like 
\href{https://huggingface.co/datasets/bjoernp/tagesschau-2018-2023}{Tagesschau} 
and the \href{https://wortschatz.uni-leipzig.de/de/download/German#deu_news_2023}{Newspaper Articles collection from the Wortschatz Leipzig University Project}. 
After evaluating these resources, we selected \href{https://huggingface.co/datasets/occiglot/occiglot-fineweb-v1.0}{occiglot-fineweb-v1.0} as the sole source for our German corpus. The dataset contains roughly 430 million cleaned documents spanning 10 languages and combines curated collections with pre-filtered web data. Using a custom German tokenizer, processing required about seven hours on a CS-2 user node and produced 110.96 billion German tokens.

\textit{Bavarian Corpus:
}
The Bavarian corpus consists of 20M tokens, assembled from the datasets listed in Table \ref{datasets-bavarian}.
\begin{table}[htbp]
\centering
\tabcolsep=0.11cm
\begin{tabular}{lll}
\toprule
\textbf{Source} & \textbf{Type} & \textbf{Size} \\
\midrule
Uni Leipzig     & Wikipedia      & 100k \\
Opus            & Wikimatrix     & 579k tokens \\
Opus            & XLEnt          & 19k tokens \\
Opus            & Wikimedia      & 55k tokens \\
Opus            & Tatoeba        & 0.6k tokens \\
Filt. FineWeb   & Wiki/Non-Wiki  & 262k sentences \\
\bottomrule
\end{tabular}
\caption{Overview of Bavarian language corpus.}
\label{datasets-bavarian}
\end{table}

\textbf{Corpus Statistics for GENBA-10B:
}
The three language-specific sources were combined into the final training corpus, with the German and English portions each truncated at 82 billion tokens and the Bavarian portion upsampled to 80 million tokens. The overall composition is summarized in Table \ref{FinalDataset}

\begin{table}[htbp]
\centering
\tabcolsep=0.11cm
\begin{tabular}{lll}
\toprule
\textbf{Language} & \textbf{Dataset} & \textbf{Tokens} \\
\midrule
English   & EnglishDataset   & 82B \\
German    & GermanDataset    & 82B \\
Bavarian  & BavarianDataset  & 80M \\
\bottomrule
\end{tabular}
\caption{Corpus composition for GENBA-10B.}
\label{FinalDataset}
\end{table}

\subsection{Tokenizer Optimization}
Existing research shows that tokenizer choice has a major impact on multilingual LLM performance, yet it often goes under-examined \cite{ali2024tokenizer}. In our English–German–Bavarian model, ensuring that German umlauts (ä, ö, ü) and the eszett (ß) are handled correctly was essential to prevent token fragmentation and preserve semantic integrity. To address this challenge, the authors systematically expanded the Llama-3-8B tokenizer vocabulary to better accommodate German and Bavarian linguistic elements through a three-phase process: 

\textbf{1. Initial Validation:} 
We ran a tokenization script on a small, high-frequency sample of Bavarian tokens drawn from our corpus. After confirming that fragmentation was minimal, we applied the same script to the full Bavarian corpus.

\textbf{2. Vocabulary Extension via Byte-Pair Encoding (BPE):)
} We merged common subword units into the original 128,256-token Llama-3 vocabulary, yielding three expanded variants:

\begin{itemize}
    \item 10\% expansion: +12,800 tokens (141,053 total), Fertility Score: 1.9026
    \item 20\% expansion: +25,600 tokens, Fertility Score: 1.8372
    \item 30\% expansion: +38,400 tokens, Fertility Score: 1.8214
\end{itemize}

\textbf{3. Performance Evaluation:} We evaluated these configurations using fertility scores, where lower scores indicate fewer tokens required per word, a desirable characteristic that reduces sequence lengths and improves training and inference efficiency. The evaluation showed a sharp fertility score decline from 10\% to 20\% expansion (1.9026 to 1.8372), but minimal improvement from 20\% to 30\% (1.8372 to 1.8214).  Balancing diminishing returns against memory and compute costs, we selected the 20 per cent expansion for our final tokenizer.

\subsection{Model Architecture Selection}
GENBA-10B employs a standard transformer-based, decoder-only architecture, following 
\citet{vaswani2017attention}. We build upon the Llama 3 \cite{dubey2024llama} architecture, which shares structural foundations with Llama 2 \cite{touvron2023llama} while offering extended context (8K versus 4K tokens) and being trained on significantly larger and more multilingual data.

\begin{figure}[h!] 
    \centering
    \includegraphics[width=0.4\textwidth]{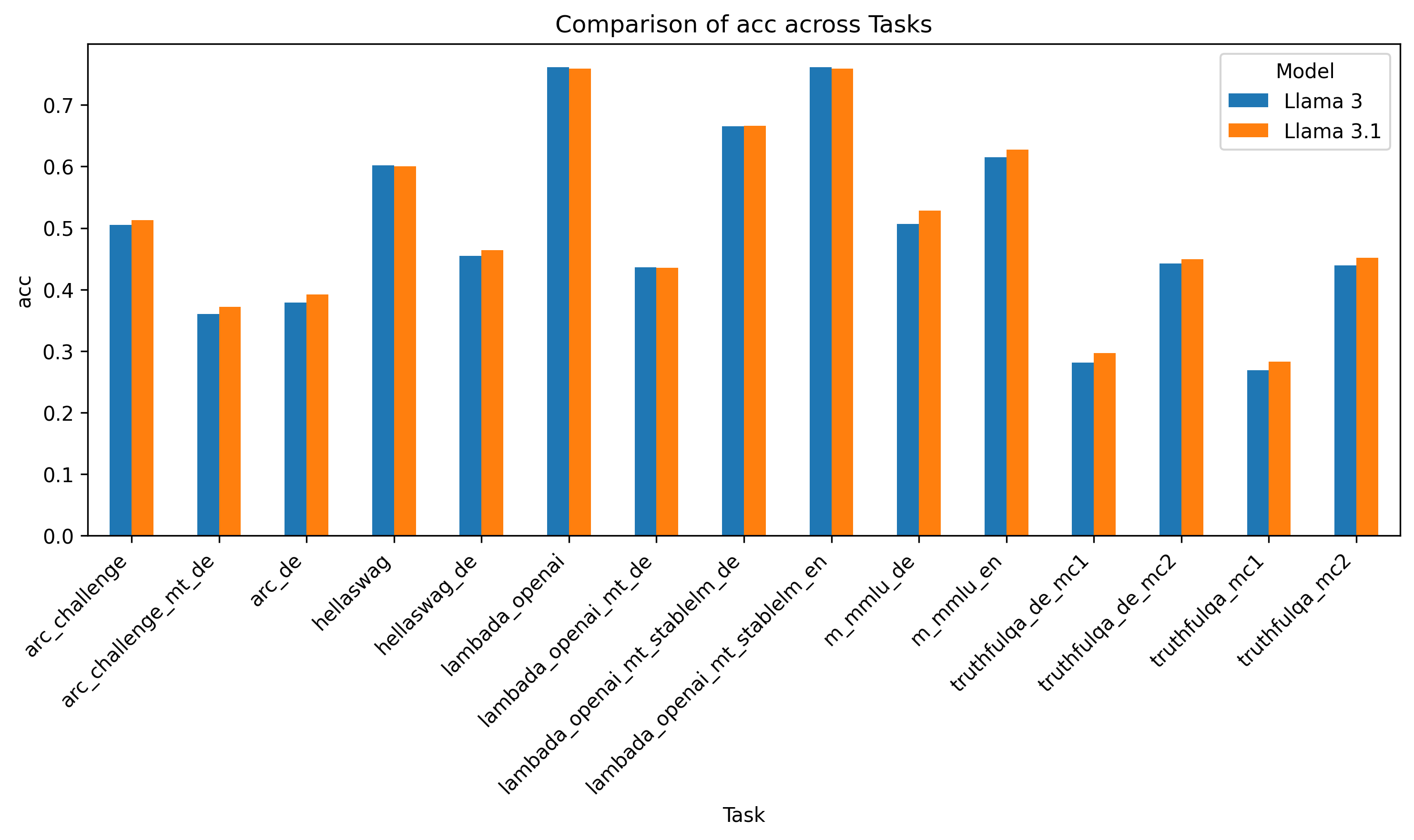}
    \caption{Comparative Performance Across Evaluation Tasks}
    \label{comparative_performance}
\end{figure}

To determine the optimal architectural foundation, we conducted comparative experiments between the Llama 3 and Llama 3.1 variants across established benchmarks. The results are presented in Figure \ref{comparative_performance}. It shows that Llama 3.1 consistently outperforms Llama 3 across multiple evaluation tasks. 

Based on these empirical results, we selected Llama 3.1 as the foundational model for subsequent training and development.

\subsection{Training Infrastructure}
Model training was conducted on the Cerebras Wafer-Scale Engine 2 (CS-2), a single-chip AI accelerator with 850\,000 AI optimized compute cores, 40 GB of on-chip SRAM, and a memory bandwidth of 20 PB/s \cite{cerebrasc2system}. The architecture also provides 220 Pb/s of fabric interconnect bandwidth, enabling high-throughput parallelism over large datasets and mitigating the scalability challenges of distributed GPU training such as inter-node communication and synchronization costs. This setup enabled efficient training of the trilingual model at scale.

\section{Experiments}
\label{Experiments}
\subsection{Finding the Right Mix of Languages}
In developing the JAIS language models, \citet{sengupta2023jais} found that mixing Arabic and English at a 1:2 ratio outperformed training on Arabic alone (p. 9). Inspired by this result, we conducted controlled experiments varying the allocation of English and German data prior to pretraining. Each experiment was constrained to a 16B token budget, with the German–English split adjusted across two scenarios. To ensure comparability, the relative proportions of the English subcorpora (Cosmopedia, Ultra Textbooks, Proof Pile 2, and Pile Knowledge) were fixed at 1:1:1:1.

 \begin{itemize} 
\item Experiment A (1:1) allocated 8B tokens to German and 8B tokens to English.
\item Experiment B (9:1) allocated 14.4B tokens to German and 1.6B tokens to English.

\end{itemize}

\begin{table}[htbp]
\centering
\tabcolsep=0.15cm
\begin{tabular}{lll}
\toprule
\textbf{Experiment} & \textbf{Acc-EN} & \textbf{Acc-DE} \\
\midrule
A (1:1)  & 0.4040 & 0.5256 \\
B (9:1)  & 0.4009 & 0.5169 \\
\bottomrule
\end{tabular}
\caption{Results for different language mixes.}
\label{results-language-mix}
\end{table}

\begin{figure}[h!] 
    \centering
    \includegraphics[width=0.4\textwidth]{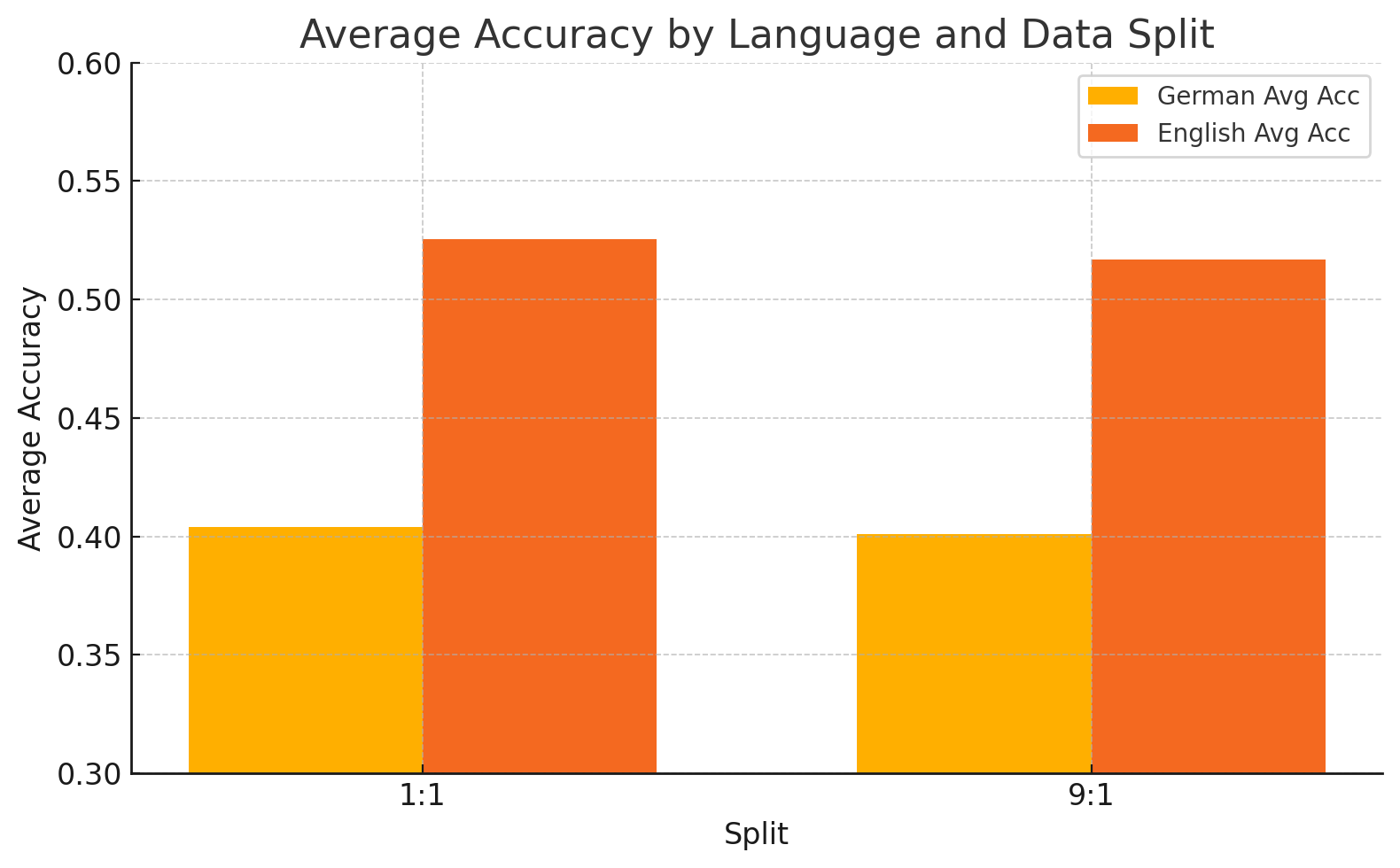}
    \caption{Performance For Different Language Splits}
    \label{ResultsLanguageMixFig}
\end{figure}

Table \ref{results-language-mix} and Figure \ref{ResultsLanguageMixFig} present accuracy results showing that a balanced 1:1 English-German ratio yields optimal performance for our trilingual model. We applied this language mix for the initial 90 per cent of pretraining before incorporating also Bavarian datasets.

\subsection{Architecture Expansion and Pretraining Methodology}
We expanded the LLaMA-3.1-8B base architecture by 20 percent through the addition of eight Transformer blocks before pretraining. In the newly inserted blocks, the FFN’s final linear layer and the attention output projection layer were initialized with zero weights, while the model-wide embedding and output layers were extended with semantically derived vectors for the new tokens. Zero-initialization, combined with backbone freezing, enables the new layers to learn meaningful transformations without altering the pretrained backbone.

Pretraining was conducted with the aim of balancing multilingual knowledge acquisition across languages of differing resource availability. For the first 90 percent of training iterations, only English and German data were used, enabling the model to establish strong representations in these high-resource languages. In the final 10 per cent, Bavarian data were introduced, upsampled and combined with the remaining English and German data to create a trilingual training regime.

This staged integration was motivated by two considerations: (i) the expanded architecture provides capacity to represent a broader multilingual feature space, and (ii) delaying the introduction of Bavarian reduces potential interference between high- and low-resource languages. Progressive incorporation of Bavarian thus allows the model to allocate capacity to this low-resource variety without immediate competition from the larger English and German datasets.

\subsection{Pretraining Hyperparameters}
GENBA-10B pretraining employed the hyperparameters detailed in Table \ref{HyperparameterTrainingSettings}. We implemented a learning rate schedule comprising an initial warmup phase where the rate linearly increased from zero to 0.00015 over approximately 1 per cent of training steps (417 steps), followed by cosine annealing decay to 0.000015 over the remaining 41,290 steps. This schedule mitigates early training instability through gradual warmup while enabling refined convergence through progressive learning rate reduction in later stages.

\begin{table}[htbp]
  \centering
  \begin{tabular}{lr}
    \toprule
    \textbf{Hyperparameter} & \textbf{Value} \\
    \midrule
    Parameters         & 10B \\
    Attention heads    & 32 \\
    Layers             & 40 \\
    Training tokens    & 164B \\
    Training steps     & 41,707 \\
    Final learning rate& $1.5\times10^{-5}$ \\
    Batch size         & 4M tokens \\
    \bottomrule
  \end{tabular}
  \caption{Pretraining hyperparameters}
  \label{HyperparameterTrainingSettings}
\end{table}
 
\subsection{Supervised Fine-Tuning}
\label{SupervisedFinetuningMethod}
We conducted supervised fine-tuning of the Llama-GENBA-10B-base model to enhance its multilingual instruction-following capabilities. In contrast to the pretraining stage, during which the Llama backbone was frozen, all model parameters remained trainable in this phase. Fine-tuning proceeded for three epochs, with checkpoints saved and evaluated at the end of each. The checkpoint from epoch 2 demonstrated the strongest performance and was therefore selected as the final fine-tuned version of GENBA-10B (Llama-GENBA-10B-instruct).

The fine-tuning process was supported by the GENBA-10B-Post-Training dataset (see Table \ref{posttrainingdataset}), 
which we constructed through systematic extension of established instruction-following corpora. 
As the foundation, we employed the \href{https://huggingface.co/datasets/argilla/databricks-dolly-15k-curated-multilingual}{Databricks-dolly-15k} dataset, 
providing high-quality English and German instruction–response pairs. To broaden multilingual coverage, these pairs were automatically translated into Bavarian using the Gemini-flash model \cite{comanici2025gemini}, 
yielding 15,000 additional pairs for each target language.

The same translation methodology was subsequently applied to the 
\href{https://huggingface.co/datasets/yahma/alpaca-cleaned}{Alpaca} and 
\href{https://huggingface.co/datasets/teknium/openhermes}{OpenHermes}
datasets, ensuring methodological consistency and comparable quality across all corpora. Hence in total, the GENBA-10B-Post-Training dataset comprises 867k instruction–response pairs with balanced representation of English, German, and Bavarian. A detailed composition and statistical analysis are presented in Table \ref{posttrainingdataset}.

\begin{table}[htbp]
  \centering

  \begin{tabular}{lrr}
    \toprule
    \textbf{Dataset} & \textbf{Per language} & \textbf{Total} \\
    \midrule
    Databricks-Dolly15k & 15k  & 45k  \\
    Alpaca              & 61k  & 183k \\
    OpenHermes          & 213k & 639k \\
    \midrule
    \textbf{Total}      & \textbf{289k} & \textbf{867k} \\
    \bottomrule
  \end{tabular}
  \caption{Datasets Used for Supervised Fine-Tuning}
  \label{posttrainingdataset}
\end{table}

\subsection{Energy Consumption Analysis}
We measured energy consumption during GENBA-10B pretraining by monitoring Power Supply Unit (PSU) measurements across the Wafer Scale Engine throughout the training process. Our measurements tracked power consumption patterns to understand the energy characteristics of wafer-scale computing during large language model training.

\section{Evaluation and Results}
\label{Results}
\begin{figure*}[htbp]
    \centering
    \includegraphics[width=\textwidth]{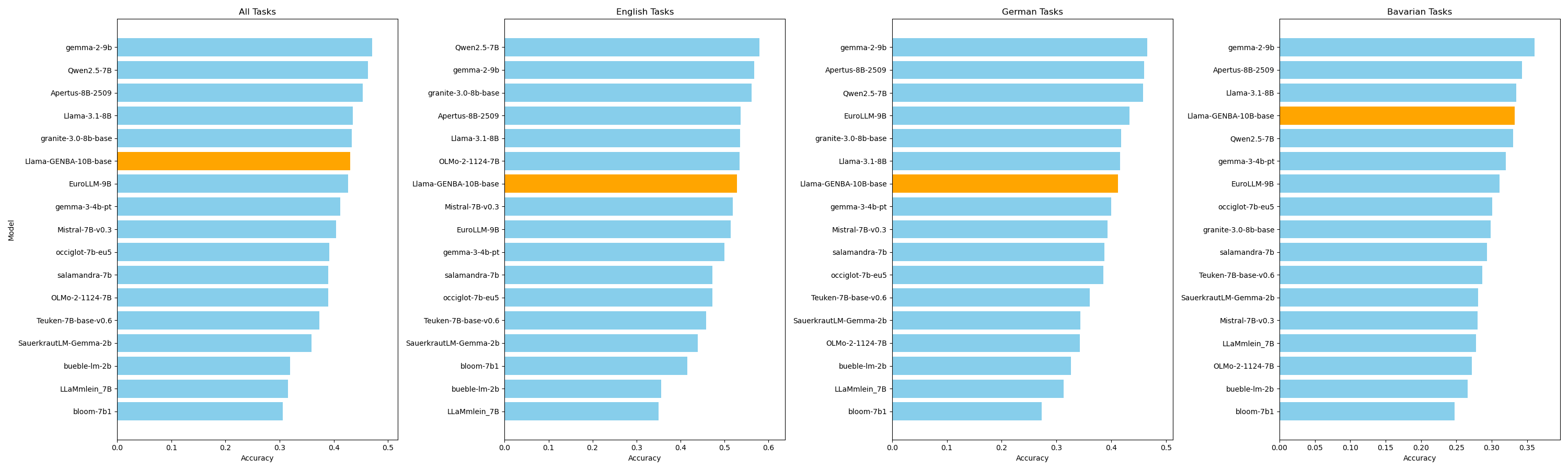}
    \caption{Comparison of Llama-GENBA-10B-base against peer base models, showing competitive performance with strong results in English, moderate results in German, and solid results in Bavarian, where it ranks fourth among baseline models.}
    \label{baseresults}
\end{figure*}

\begin{table*}[htbp]
\centering
\setlength{\tabcolsep}{4pt}
\renewcommand{\arraystretch}{1.2}
\scriptsize

\begin{adjustbox}{max width=\textwidth}
\begin{tabular}{lccccc}
\toprule
\textbf{Pre-trained} & Arc & Hellaswag & MMLU & TruthfulQA & Winogrande \\
\midrule
\multicolumn{6}{l}{\textit{Non-European}} \\
\midrule
Llama-3.1-8B              & 0.3840 & 0.4679 & 0.5322 & 0.3528 & 0.7380 \\
OLMo-2-1124-7B            & 0.3256 & 0.4213 & 0.4362 & 0.3331 & \textbf{0.7459} \\
Qwen2.5-7B                & 0.3769 & 0.4497 & \textbf{0.6229} & \textbf{0.4043} & 0.7269 \\
SauerkrautLM-Gemma-2b     & 0.3034 & 0.4016 & 0.3484 & 0.3262 & 0.6811 \\
gemma-2-9b                & \textbf{0.4750} & \textbf{0.5032} & 0.5992 & 0.3425 & 0.7411 \\
gemma-3-4b-pt             & 0.3888 & 0.4526 & 0.4893 & 0.3213 & 0.6914 \\
granite-3.0-8b-base       & 0.3696 & 0.4737 & 0.4702 & 0.3852 & 0.7443 \\
\midrule

\multicolumn{6}{l}{\textit{European}} \\
\midrule
Apertus-8B-2509           & \textbf{0.4235} & \textbf{0.4949} & \textbf{0.5107} & \textbf{0.3831} & 0.6890 \\
EuroLLM-9B                & 0.3771 & 0.4718 & 0.4615 & 0.3737 & 0.6946 \\
LLaMmlein\_7B             & 0.2593 & 0.3951 & 0.2359 & 0.3104 & 0.5722 \\
Mistral-7B-v0.3           & 0.3585 & 0.4483 & 0.4428 & 0.3382 & 0.7348 \\
Teuken-7B-base-v0.6       & 0.3459 & 0.4528 & 0.3910 & 0.3140 & 0.6914 \\
bloom-7b1                 & 0.2282 & 0.3449 & 0.2500 & 0.3088 & 0.6440 \\
bueble-lm-2b              & 0.2482 & 0.3401 & 0.2536 & 0.3514 & 0.5454 \\
occiglot-7b-eu5           & 0.3545 & 0.4658 & 0.3861 & 0.3317 & 0.6930 \\
salamandra-7b             & 0.3754 & 0.4605 & 0.3494 & 0.3351 & 0.6875 \\
\textbf{Llama-GENBA-10B-base}            & 0.3776 & 0.4792 & 0.4631 & 0.3729 & \textbf{0.7364} \\
\bottomrule
\end{tabular}
\end{adjustbox}

\caption{Comparison of Llama-GENBA-10B-base model versus other LLMs on core benchmarks (ARC, HellaSwag, MMLU, TruthfulQA, Winogrande). Llama-GENBA-10B-base shows balanced performance across tasks, with particularly strong Winogrande results nearly matching top-performing systems.} 
\label{baseResultsTable}
\end{table*}

We adopt a three-part evaluation strategy: (i) evaluation of the Llama-GENBA-10B-base model against comparably sized systems, (ii) evaluation of the supervised fine-tuned Llama-GENBA-10B-instruct model, and (iii) analysis of GENBA-10B’s (Llama-GENBA-10B-base) energy consumption during pretraining.

\subsection{Evaluation of the Base Model}

\textbf{Benchmarks:}
\label{BenchmarksForAutoEvaluation}
Evaluation was conducted using the EleutherAI evaluation harness, including both the original English version \cite{eval-harness} and its  \href{https://github.com/bjoernpl/GermanBenchmark?tab=readme-ov-file}{German} \cite{dac2023okapi} adaptation. For Bavarian we translated the German versions of these benchmarks to enable evaluation in

\begin{figure*}[htbp]
    \centering
    \includegraphics[width=\textwidth]{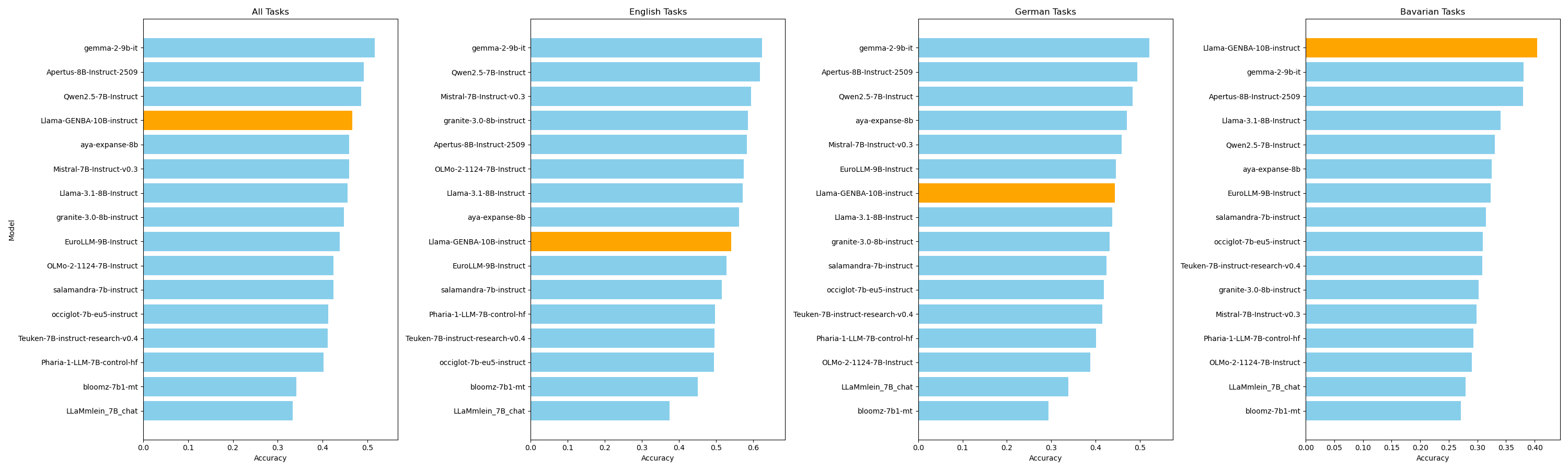}
    \caption{Performance comparison of Llama-GENBA-10B-instruct against peer instruction-tuned models, showing GENBA-10B-instruct as the best-performing model in Bavarian.}
    \label{comparisionEurope}
\end{figure*}

To measure reasoning capabilities, we employ ARC-Challenge \cite{clark2018think}, which consists of 7,787 grade-school multiple-choice science questions.

For commonsense reasoning, we use HellaSwag \cite{zellers2019hellaswag}, a benchmark requiring models to select the most plausible continuation of a partial event description; the English set contains 59,950 instances.

We also include TruthfulQA \cite{lin2021truthfulqa}, an 817-question benchmark covering 38 categories such as health, law, finance, and politics, which evaluates whether models generate factually accurate answers.

To assess broad, domain-diverse knowledge and reasoning, we use MMLU \cite{hendrycks2020measuring}, comprising 15,908 multiple-choice questions from 57 subjects spanning STEM, humanities, social sciences, and professional domains.

Finally, we use WinoGrande \cite{winogrande}, a large-scale pronoun resolution benchmark with roughly 44,000 fill-in-the-blank problems inspired by the Winograd Schema Challenge but expanded in size and difficulty.

\textbf{Results:} 
Figure \ref{baseresults} presents a comparative evaluation of Llama-GENBA-10B-base against other LLMs of similar scale. Overall, Llama-GENBA-10B-base trails models such as gemma-2-9b, Qwen2.5-7B, Apertus-8B-2509, Llama-3.1-8B, and granite-3.0-8B-base. However, it outperforms several other models, particularly those developed in Europe, including EuroLLM-9B, Teuken-7B-base-v0.6, SauerkrautLM-Gemma-2B, bueble-lm-2b, LLaMmlein-7B, and bloom-7B1. The reported aggregated accuracies are derived from individual task-specific results (HellaSwag, MMLU, TruthfulQA, and Winogrande), with further details provided in Table \ref{baseResultsTable} .

On English benchmarks, Llama-GENBA-10B-base performs comparatively strongly, placing close to the top-performing models. It lags only slightly behind Qwen2.5-7B, gemma-2-9B, Granite-3.0-8B,  Apertus-8B-2509, Llama3.1-8B and Olmo-2-1124-7B but achieves higher accuracy than Mistral-7B-v0.3, EuroLLM-9B, and other European-developed baselines. This suggests that Llama-GENBA-10B-base has a solid foundation in English, though it is not yet at the level of state-of-the-art 7B–9B models.

In German, Llama-GENBA-10B-base performs competitively, with accuracy levels close to those of mid-tier models. It trails gemma-2-9B, Apertus-8B, Qwen2.5-7B, and EuroLLM-9B, but surpasses several European-built models such as Teuken-7B-base-v0.6, SauerkrautLLM-Gemma-2B, and bueble-lm-2b.

On Bavarian, Llama-GENBA-10B-base performs relatively well, ranking among the top models. It achieves accuracy close to leading systems such as gemma-2-9b, Apertus-8B-2509, and Llama-3.1-8B, while surpassing Qwen2.5-7B and EuroLLM-9B. This result suggests that Llama-GENBA-10B-base generalizes effectively to this dialect, even outperforming models such as EuroLLM-9B or Teuken-7B-base-v0.6.

As a key take away, Llama-GENBA-10B-base shows balanced performance across languages, with strong competitiveness in English, moderate ability in German, and surprisingly strong generalization in Bavarian. Its relative success on Bavarian suggests robustness in handling low-resource settings, which may stem from its training data composition or design choices.

\subsection{Evaluation of the Fine-Tuned Model}
After evaluating Llama-GENBA-10B-base's performance across pretraining, we compared the model's final fine-tuned version (as described in Section \ref{SupervisedFinetuningMethod}) with other models using the same benchmark datasets as for the pretraining benchmarking. As shown in Figure \ref{comparisionEurope}, fine-tuning substantially improves the model's performance across benchmarks. Compared to its base version, the instruction-tuned variant (Llama-GENBA-10B-Instruct) achieves stronger aggregate accuracy, placing close to leading mid-sized models such as gemma-2-9B-it and Qwen2.5-7B-Instruct. In English and German tasks, Llama-GENBA-10B-Instruct secures a mid-tier position while it trails top performers like gemma-2-9B-it, Apertus-8B-Instruct-2509, and Qwen2.5-7B-Instruct, it consistently outperforms several instruction-tuned baselines, including salamandra-7B-instruct and Teuken-7B-instruct-research-v0.4. 

The most striking gain is observed on Bavarian tasks. Whereas the base Llama-GENBA-10B-base ranked among the top five models, Llama-GENBA-10B-instruct advances to the top position, outperforming all evaluated systems, including gemma-2-9B-it and Apertus-8B-Instruct-2509. This shift underscores the effectiveness of instruction tuning for a low-resource and dialectal setting, where general-purpose pretraining alone is insufficient. 

Together, these results demonstrate that Llama-GENBA-10B-instruct benefits substantially from instruction tuning, with the fine-tuned variant surpassing Apertus-8B-Instruct-2509 and gemma-2-9B-it in Bavarian and establishing itself as the best model in its class for this language, while also outperforming EuroLLM-9B-instruct in English and matching its results in German.

\subsection{Energy Usage of Pre-Training}
The pretraining of the Llama-GENBA-10B-base model consumed approximately 35.23 megawatt-hours (MWh) of electricity over a period of 66 days, using the Cerebras CS-2 system (See Table \ref{energy-comparisons}). This amount of energy is roughly equivalent to the annual electricity usage of ten average European households, based on 2022 Eurostat data \cite{eurostat2022}\footnote{Based on 2022 Eurostat data, average EU household electricity use was about 3.6 MWh per year.}.

\begin{table}[htbp]
\centering
\tabcolsep=0.15cm
\begin{tabular}{ll}
\toprule
\textbf{Metric} & \textbf{CS-2} \\
\midrule
Average power draw        & 22.3 kW \\
Training duration         & 66 days \\
Total energy consumption  & 35.23 MWh \\
\bottomrule
\end{tabular}
\caption{Estimated energy usage during pretraining.}
\label{energy-comparisons}
\end{table}

\section{Conclusion and Future Work}
\label{Conclusion}
We present Llama-GENBA-10B, a trilingual foundation model comprising both a base version (Llama-GENBA-10B-base) and an instruction-tuned variant (LLama-GENBA-10B-instruct). Built on Llama3.1-8B and scaled to 10B parameters, GENBA-10B is pretrained on 164B tokens distributed evenly between English and German, complemented by 80M tokens of Bavarian. This design mitigates English dominance, strengthens resources for the German NLP community, and promotes Bavarian as a low-resource language. Model development addressed four central challenges: (i) constructing a balanced multilingual corpus, (ii) designing a unified tokenizer, (iii) optimizing architectural and data ratios for cross-lingual transfer, and (iv) creating the first trilingual evaluation suite by translating German benchmarks into Bavarian.

Our evaluation demonstrates that the Llama-GENBA-10B-base model achieves competitive performance, with strong results in English, moderate outcomes in German, and state-of-the-art performance in Bavarian. The instruction-tuned variant further improves upon this, surpassing both Apertus-8B-instruct-2509 and gemma-2-9B-it in Bavarian, thereby establishing a new benchmark for this language. In addition, it outperforms EuroLLM-9B in English and reaches parity in German.

Furthermore, our experiments on the Cerebras CS-2 confirm that large-scale multilingual pretraining is feasible for small research teams, with the added benefit of real-time energy monitoring. Taken together, Llama-GENBA-10B provides a blueprint for building linguistically inclusive and resource-efficient foundation models.

Future work could focus on instruction-tuning safety through harm risk categorization and detection mechanisms. We may integrate this model into a chatbot for manual evaluation and conduct systematic human assessment of outputs across all three languages. More broadly, this multilingual LLM development approach on the Cerebras CS-2 System could be applied to other dialects and endangered languages, potentially enabling new downstream applications.

\bibliography{mybibfile}
\bibliographystyle{acl_natbib}

\appendix

\end{document}